\documentclass[times,twocolumn,final,authoryear]{elsarticle}

\usepackage{framed,multirow}

\usepackage{amssymb}
\usepackage{latexsym}

\usepackage{times}
\usepackage{epsfig}
\usepackage{graphicx}
\usepackage{subcaption}
\usepackage{amsmath}
\usepackage{multirow}
\usepackage{adjustbox}
\usepackage{hhline}
\usepackage{lipsum}

\usepackage{url}
\usepackage{xcolor}
\definecolor{newcolor}{rgb}{.8,.349,.1}
\newcommand{\R}{\mathbb{R}}
\journal{Pattern Recognition Letters}

\begin{document}

\begin{frontmatter}

\title{Hierarchical Attention Network for Action Segmentation}

%
%

\author[label1]{Harshala~Gammulle\corref{cor1}}
\ead{pranali.gammulle@hdr.qut.edu.au}

 \author[label1]{Simon~Denman}
\ead{s.denman@qut.edu.au}
 
\author[label1]{Sridha~Sridharan}
\ead{s.sridharan@qut.edu.au}

 \author[label1]{ Clinton~Fookes}
\ead{c.fookes@qut.edu.au}

 \cortext[cor1]{Corresponding author at: Image and Video Research Laboratory, SAIVT, Queensland University of Technology, Australia.}
 \address[label1]{Image and Video Research Laboratory, SAIVT, Queensland University of Technology, Australia.}

\begin{abstract}
Temporal segmentation of events is an essential task and a precursor for the automatic recognition of human actions in the video. 
Several attempts have been made to capture frame-level salient aspects through attention but they lack the capacity to effectively map the temporal relationships in between the frames as they only capture a limited span of temporal dependencies. To this end we propose a complete end-to-end supervised learning approach that can better learn relationships between actions over time, thus improving the overall segmentation performance. The proposed hierarchical recurrent attention framework analyses the input video at multiple temporal scales, to form embeddings at frame level and segment level, and perform fine-grained action segmentation. This generates a simple, lightweight, yet extremely effective architecture for segmenting continuous video streams and has multiple application domains. We evaluate our system on multiple challenging public benchmark datasets, including MERL Shopping, 50 salads, and Georgia Tech Egocentric datasets and achieves state-of-the-art performance. The evaluated datasets encompass numerous video capture settings which are inclusive of static overhead camera views and dynamic, ego-centric head-mounted camera views, demonstrating the direct applicability of the proposed framework in a variety of settings. 
\end{abstract}

\begin{keyword}

\end{keyword}

\end{frontmatter}

\section{Introduction}

Actions performed by humans are continuous in nature, and different actions flow seamlessly from one to another as a person completes one or more tasks. Hence when designing intelligent surveillance systems it is crucial to utilise robust and efficient action segmentation mechanisms to recognise human actions from those continuous streams. Such a system is not only beneficial for detecting suspicious activities in a surveillance setting but also vital for numerous tasks including elder care monitoring, human-robot interactions and warehouse automation. 

Action segmentation algorithms should exploit the idea that individual actions in a continuous sequence are often related to its surrounding action. For example, when performing an activity such as shopping, the action `inspect item on shelf' may be followed by the actions `put item in basket'. However, it may also witness a `theft' event. Each action is related and dependant on what has previously happened, as such, capturing the relations among those actions facilitates the segmentation process. 

Most existing approaches for continuous action segmentation \citep{lea2016,merlshopping} use features extracted from Convolution Neural Networks (CNNs), at each  frame, and map them temporally through Recurrent Neural Network (RNN) models. However such naive temporal modelling has been shown to capture a limited span of temporal dependencies \citep{merlshopping}. Temporal Convolution Networks (TCNs) \citep{leaCVPR} have been introduced to capture attention at a frame level and are further augmented in \cite{ding2017tricornet} through a hybrid of both TCN and RNN to model dependencies at multiple temporal scales. However one obvious drawback of this architecture is that it has the overhead of learning the temporal attention values of the TCN kernels as well as the RNN parameters for long-term dependencies. 

\begin{figure}[h!]
        \centering
        	\includegraphics[width=0.97\linewidth]{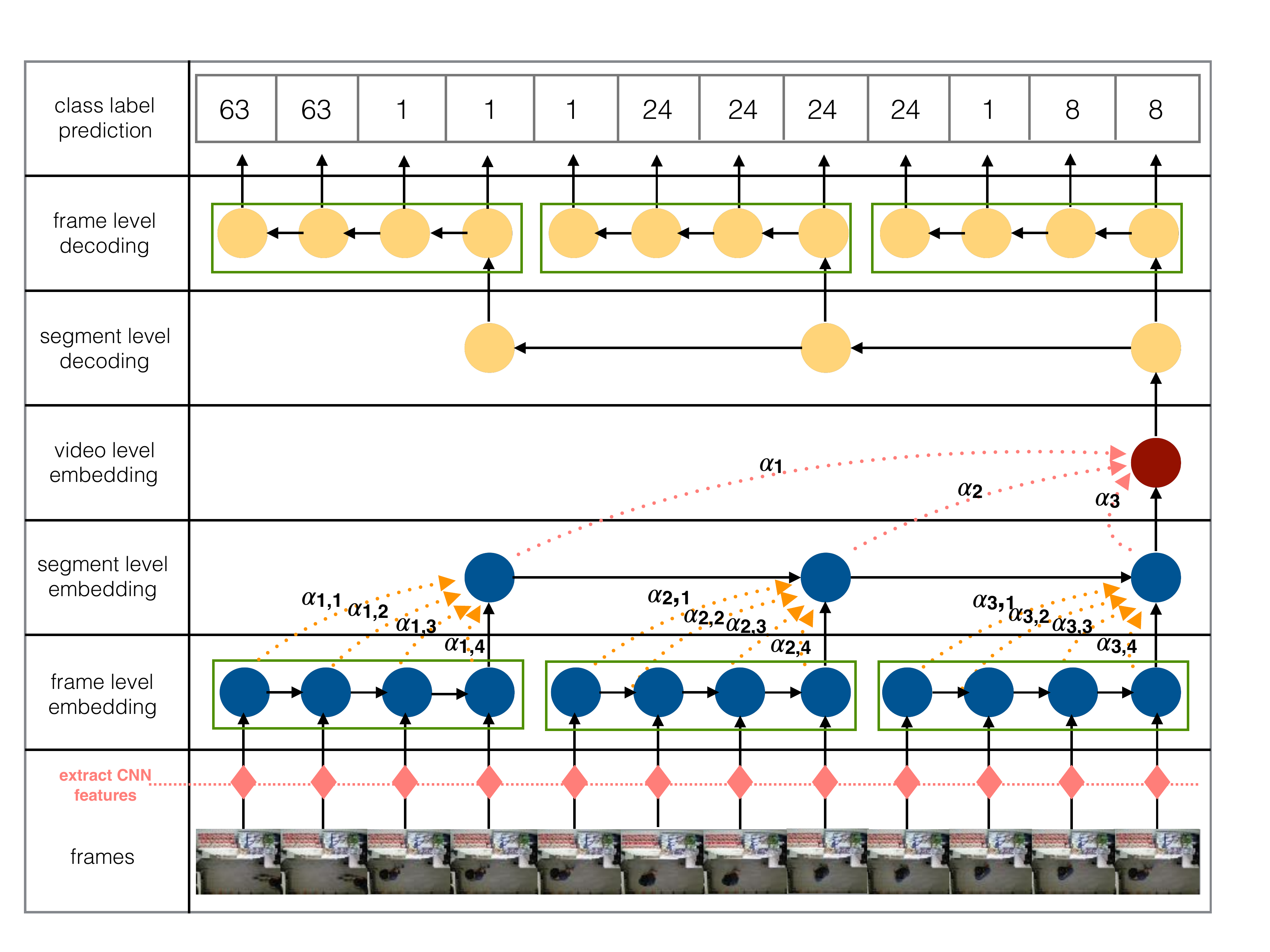}
	\caption{Proposed hierarchical attention framework: Each input frame passes through a CNN to extract out salient spatial features. Then the temporal relationships at a frame level are captured in the first recurrent layer of the encoder, utilising the frame level attention weights $\alpha_{i,t}$ (where $i$- number of segments and $t$- number of frames in each segment) to generate a single vector to represent the context of the video segment. At the next stage we map the temporal relationships between different segments of the input video using a second recurrent layer and these embeddings are combined and compressed using a series of segment level attention weights $\alpha_{i}$ in order to generate a unique representation for the entire video. During the decoding stage these representations are expanded at each level of the hierarchy and finally mapped to the respective class label through the frame level decodings. }
	\label{fig:model_arc}
\end{figure}

In contrast we introduce a simpler, yet effective hierarchical attention framework which learns both frame level and segment level attention parameters in an unsupervised manner. The proposed framework is shown in Fig. \ref{fig:model_arc}. The hierarchical network architecture is motivated by \cite{hierLSTM} and \cite{sordoni2015hierarchical} and shows the ability to learn the temporal correspondences of the extracted CNN spatial features while learning relationships between actions that facilitate the overall action segmentation task. 

As shown by Fig.\ref{fig:model_arc}, we perform frame level embedding on sub-segments of the video at the first level of the hierarchy and utilise frame attention weights $\alpha_{i,t}$ (where $i$- number of segments and $t$- number of frames in each segment), to retain only significant information when deriving the segment level embeddings. At the next stage, these segments are combined and compressed using a series of segment attention weights $\alpha_{i}$, in order to generate a representation for each video sequence. At the decoding stage, we expand these representations at each level of the hierarchy and finally at the frame level decoding we map the representation to a class label. Hence, our end-to-end model maps the sequence of video frame features directly to class labels and is able to compete with the current state-of-the-art models.

Our main contributions are listed as follow:

\begin{itemize}
  \item We propose a novel hierarchical recurrent model that performs end-to-end continuous fine-grained action segmentation.
  \item We introduce an attention framework that effectively embeds salient frame level and video segment level spatio-temporal features to facilitate the overall action segmentation task.
  \item We achieve state-of-the art results for three challenging datasets: MERL shopping, 50 salads, Georgia Tech Egocentric dataset.
  \item We conduct a series of ablation experiments to demonstrate the importance of different components of the proposed hierarchical model.
  
\end{itemize}

Due to the lack of availability of annotated data from real world surveillance videos, we utilise the MERL Shopping \citep{merlshopping}, 50 salads \citep{stein2013combining}, and Georgia Tech Egocentric \citep{li2015delving} datasets in our evaluations. However with the inclusion of MERL Shopping dataset which mimics a surveillance setting, captured through an overhead camera; the dynamic egocentric camera view provided through a head-mounted camera in the Georgia Tech Egocentric dataset and the home monitoring system instantiated in the 50 salads dataset, our evaluations suggest the direct application of the proposed architecture in any surveillance setting with either static or dynamic camera views is possible. 

\section{Related work} 
\label{related_work}

Action recognition can be broadly grouped into two main categories: Discrete \citep{imageBased1, twoStream, gammulle2017two, gammulle2019predicting} and continuous fine-grained action recognition  \citep{leaCVPR, Ni2014, zhou2014, zhou2015, gammulle2020fine, gammulle2019forecasting} approaches. Discrete action recognition is performed on pre-segmented video sequences containing one action per video while continuous fine-grained models are evaluated on untrimmed video sequences containing more than one action per video. Since actions occurring in the real world are continuous in nature finding solutions to continuous fine-grained action recognition is more important and is also more challenging as it requires both localisation and recognition of the action frames. We therefore focus on continuous fine grained action scenarios in this paper.

In early literature on continuous action recognition, most works utilise hand-crafted features such as HOG, HOF, MBH, Trajectory or pose based features for the action segmentation task  \citep{videoDarwin,cooking1}. Apart from these, human-object interactions have been considered in \citep{de2008, trinh2012} by  taking hand-object related features into account. However, these methods lack the ability to capture rich contextual information among various objects during interactions. \cite{zhou2014} introduced a model to overcome these limitations, capturing joint distributions between motion and objects-in-use informations.

Most recently with the introduction of deep learning, researchers have focused on utilising deep neural networks for feature extraction and temporal mapping. For instance, \cite{merlshopping} utilise multiple Convolution Neural Networks (CNNs) for feature extraction from multiple input feature streams at the frame level and map their temporal dependencies through a bi-directional Recurrent Neural Network (RNN). \cite{wu2018deep} investigated Convolutional RBMs for feature extraction from video frames. In \cite{wu2018and,wu2018deep2} the authors propose spatially recurrent architectures for deep feature embedding and \cite{wu2018and2} investigated Gated Recurrent Unit (GRUs) coupled with location-based attention as a method for feature extraction. However, \cite{leaCVPR} demonstrate the limited attention span of native RNN architectures and several architectures have been proposed using Temporal Convolution Networks (TCN) \citep{leaCVPR,lea2016temporal} for generating the required frame level attention, and they are further augmented in \cite{ding2017tricornet} with a hybrid of TCN and RNN to overcome the fixed temporal attention span of the TCN model.  

However we observe several draw backs of these architectures. Firstly the TCN attention mechanism considers fixed, short temporal windows when embedding the input frames. Furthermore, the introduction of bi-directional RNN in \cite{ding2017tricornet} to map long-term temporal relationships inherits the additional overhead of training both TCN and RNN models to propagate the frame level information to the decoder model. In addition, several prior works have shown naive RNN architecture without an attention mechanism to loose long-term relationships and focus only on the last few time-steps \citep{kumar2016ask, fernando2018tree}. In contrast, we propose a simple and light weight approach using a hierarchical attention framework to encode the salient spatio-temporal information effectively. We capture our inspiration from the natural language processing domain where they illustrate the efficiency of hierarchical representation learning when handling lengthier sequences. 


Hierarchical learning with time series data has been extensively investigated as it provides a richer context representation compared to standard unstratified learning \citep{deng2016deep}. For instance in \cite{deng2016deep}, the authors investigated the hierarchical learning paradigm by coupling deep learning together with reinforcement learning to model human behaviour in financial asset trading. They illustrated that a deep recurrent neural network front end for their proposed framework is more effective for learning informative features from time series data. A similar observation is reported in \cite{deng2016hierarchical}, where the authors propose to hierarchically encode the information derived from both neural and fuzzy system to generate a richer embedding. The effectiveness of this approach is validated on different machine learning tasks including image categorisation, segmentation and financial data prediction.

 
When considering image based works, \cite{you2014local}  proposed to use a hierarchical metric learning approach for object detection. An approach which simultaneously learns a distance metric for inter-video and intra-video distances, which could also be seen as hierarchical representation learning from videos, is proposed in  \cite{zhu2018video}. In a different line of work, the authors in \cite{yu2018hierarchical} have looked at pooling features from multiple level of the hierarchy to create an informative feature embedding. However, none of the related works on video based recognition have looked into utilising a hierarchical attention framework when extracting these informative features from frame embeddings. To the best of our knowledge, only a single-level of attention  \cite{yang2017deep} is utilised for most computer vision tasks. Furthermore, our model does not inherent any deficiencies of image based methods as in  \cite{deng2013low}, and is capable of handling long video sequences of RGB frames without losing information.  

The utilisation of the hierarchical attention allows us not only to achieve better efficiency compared to \cite{ding2017tricornet}, but also to gain comparatively better results on standard benchmarks. 
 
\section{Methodology} 
\label{methods}

Following \citep{leaCVPR} we formulate the action segmentation task as  a multi-class classification problem, where we generate an action classification for every frame in a video. Motivated by the impressive results from \cite{hierLSTM} and \cite{sordoni2015hierarchical} for embedding long term temporal information in text synthesis and language modelling tasks, we develop a new hierarchical attention framework as an encoder-decoder framework capturing spatio-temporal information in the input video. We model the dependencies between frames at multiple levels, namely frame-level and segment-level, through the proposed hierarchical attention framework as follows. 

\subsection{Frame-level embedding}

The frame-level embedding is performed by first extracting salient spatial features from a CNN model. Let the video sequence ($X$) contain $i$ ($i \in [1 , N]$) segments, each containing $t$ ( $t \in [1 , T]$) frames. Then for each frame $x_{i,t}$ in video $X$ ($X=$\{$x_{1,1},x_{1,2},\dots,x_{N,T}$\}) CNN features ($e_{i,t}$) are extracted as follows, 

\begin{equation}
e_{i,t}=CNN(x_{i,t}).
\label{eq:2_1}
\end{equation}

The extracted frame-wise CNN features are then passed through the first encoder LSTM which generates an embedding output for each segment. 

\begin{equation}
h^{encode}_{i,t}=LSTM(e_{i,t}, h^{encode}_{i,t-1}),
\label{eq:2_2}
\end{equation}
where $h^{encode}_{i,t} \in \R^{L \times 1}$.

Then we apply frame level attention to capture different levels of salient action features from different frames.

\begin{equation}
u^{encode}_{i,t}=tanh(\hat W h^{encode}_{i,t}+ \hat b),
\label{eq:2_3}
\end{equation}

where $\hat W (\in \R^{L \times 1})$ and $\hat b$ are weight and the bias of a multi layer perceptron (MLP), and $u^{encode}_{i,t}$ is the hidden representation of the one layer MLP for frame embedding $ h^{encode}_{i,t}$. Then the similarity of $u^{encode}_{i,t}$ with the frame level context vector $\hat u$ is measured and normalised through a softmax function given by, 

\begin{equation}
\alpha_{i,t}=\frac{exp([u^{encode}_{i,t}]^T, \hat u)}{\sum_{j=1}^{T}exp([u^{encode}_{i,j}]^T, \hat u)} .
\label{eq:2_4}
\end{equation}
The frame level context vector can be seen as a higher level representation of the context of those frames. $ \hat u$ is randomly initialised and jointly trained during the training process. These weighted embeddings are then used to generate the segment embedding of the $i_{th}$ video segment as follows,

\begin{equation}
s_i= \sum_{j=1}^{T} \alpha_{i,j} h^{encode}_{i,j}.
\label{eq:2_5}
\end{equation}

\subsection{Segment-level embedding}
\label{sec:segment_level}
Given the segment embedding, $s_i$, video embedding can be performed as follows,

\begin{equation}
h^{encode}_i=LSTM(s_{i},h^{encode}_{i-1}).
\label{eq:2_6}
\end{equation}
 
 Different video segments will have different levels of visual clues when recognising the video actions. Therefore similar to, Equations \ref{eq:2_4} and \ref{eq:2_5}, we quantify the importance of each segment using attention,

\begin{equation}
u^{encode}_{i}=tanh(\dot{W} h^{encode}_{i}+\dot{b}),
\label{eq:2_7}
\end{equation}

\begin{equation}
\alpha_{i}=\frac{exp([u^{encode}_{i}]^T, \dot{u})}{\sum_{j=1}^{T}exp([u^{encode}_{j}]^T, \dot{u})} ,
\label{eq:2_8}
\end{equation}

\begin{equation}
v= \sum_{j=1}^{T} \alpha_{j} h^{encode}_j,
\label{eq:2_9}
\end{equation}

where $\dot{u}$ is the segment level context vector and $v$ is the embedding that summarises the overall video sequence, $X$.

\subsection{Decoders}

After generating a video level embedding, representing the context of the entire input video, we need to decode this information to generate the frame level class labels, segmenting the video into action classes. Similar to the encoding process we attend to the decoding in two levels, segment level decoding and frame level decoding. First we initialise the initial hidden state $h^{decode}_{0}$ of the segment level decoder using the video embedding $v$ (i.e $h^{decode}_{0} = v$) and generate the segment level decoding as follows,

\begin{equation}
h^{decode}_{i}=LSTM(s_i,h^{decode}_{i-1}).
\label{eq:3_1}
\end{equation}

Then at the next level frame level decoding is done through,

\begin{equation}
h^{decode}_{i,t}=LSTM(e_{i, t},h^{decode}_{i,t-1}),
\label{eq:3_2}
\end{equation}

where $h^{decode}_{i,0} = h^{decode}_{N} $. Finally, we generate the class label for each frame using, 
\begin{equation}
P=softmax(h^{decode}_{i,t}).
\label{eq:3_2}
\end{equation}

\section{Experimental Results}
\label{ex-results}
\subsection{Datasets}

The evaluations have been performed by utilising the following fine-grained continuous action video datasets, containing multiple actions within each video sequence.

\subsubsection{MERL Shopping dataset \citep{merlshopping}}

The MERL Shopping dataset consists of 96 videos, each with a duration of 2 minutes. The videos have been recorded by a static over head video camera, and show 32 different subjects shopping from grocery store shelving units. These videos includes five different action classes: `Reach to Shelf', `Retract from Shelf',`Hand in Shelf', `Inspect Product' and `Inspect Shelf'. Training, testing and validation splits are as in \cite{merlshopping}

\subsubsection{The University of Dundee 50 salads dataset \citep{stein2013combining}}

50 salads dataset contains 25 users, each appearing two times making a salad. These 50 videos are captured through a static RGBD camera pointed at the user and last up to 10 minutes. 
Along with the visual and depth information the authors have captured accelerometer data, however, only the video data is used for evaluations. We considered all seventeen mid-level action classes together with the background class. Training, testing and validation splits are as in \cite{stein2013combining}.

\subsection{Georgia Tech Egocentric Activities Dataset \citep{li2015delving}}

In contrast to the static overhead camera view of the previous 2 datasets, this  dataset contains videos recorded from a head mounted camera and includes four subjects performing seven different daily activities. We utilise 11 action classes defined in \cite{li2015delving} including the background class. 

\subsection{Evaluation Metrics}

The evaluations are performed utilising both segmentation and frame-wise metrics. Even though the frame-wise metric has been widely used, due to different segmentation behaviour, models with similar frame-wise accuracies can often show high visual dissimilarities \citep{leaCVPR}. As such we also utilise segmentation metrics such as segmental F1 score (F1@k) (\cite{leaCVPR}), segmental edit score (edit) (\cite{lea2016learning}) and mean average precision with midpoint hit criterion (mAP@mid) \citep{merlshopping,rohrbach2016}. Depending on the availability of the baseline model results we use different combinations of these metrics when evaluating different datasets. 

\subsection{Implementation details}

For CNN feature extraction we utilised ResNet101 \citep{resnet}, which has been pre trained on the ImageNet database \citep{imageNet}, and extracted 2048 dimensional feature vectors from the final average pooling layer per frame. These features are then passed through a fully connected layer with 200 hidden units to reduce the input feature dimension to 200. 

At the first stage of the encoding we sequentially pass input features through the first layer of the encoder with 200 hidden units in order to generate frame level embedding (see Fig. \ref{fig:model_arc}). In the next level of the hierarchy these embeddings are combined and compressed to represent the segment level representation with 200 hidden units. In the final level of the encoder we represent the entire video with 200 units. For each level in the decoder we utilise the same hidden unit dimensions as the encoder. The proposed HA\_Net model is implemented with the Keras \citep{chollet2015keras} deep learning library with Theano \citep{bergstra2010theano} backend. The network is trained using categorical cross-entropy loss with stochastic gradient decent and the Adam optimiser \citep{adam}, and initialised using default Keras settings.

\subsection{Results}

The evaluation results for MERL Shopping \citep{merlshopping}, 50 salads \citep{stein2013combining} and Georgia Tech Egocentric \citep{li2015delving} datasets are presented in Tables \ref{tab:tab_1}, \ref{tab:tab_2} and \ref{tab:tab_3}, respectively.

For MERL Shopping, as the first baseline we use the method introduced by \citep{merlshopping}, for which we use the re-evaluated results provided in \cite{leaCVPR} for frame-wise and segmentation metrics. `MSN Det' denotes the results for the sparse model proposed in \cite{merlshopping} where as `MSN Seg' denotes the dense (per frame) action segmentation model. 
 
 When observing the results presented in Table \ref{tab:tab_1} we observe poor performance from MSN Det and MSN Seg models, especially among F1 scores, mainly due to the over segmentation of the input video. This is largely because the long-term temporal dependencies cannot be captured through a naive recurrent structure such as a bi-directional LSTM. The model ED-TCN \citep{leaCVPR} has been able to overcome these deficiencies by incorporating Temporal Convolutional Networks (TCN) where they utilise multiple attention scales to capture salient spatio-temporal information.  
 
However, when observing the results presented in Tables \ref{tab:tab_2} and \ref{tab:tab_3} it is clear that the ED-TCN model doesn't fully capture the temporal context presented in the video segment. The model fails to capture long-term dependencies among different action classes in the given video due to its fixed receptive field size. TricorNet \citep{ding2017tricornet} tackles this issue through a hybrid of TCN and RNN. In order to capture long-term dependencies they pass the TCN embeddings through a bi-directional LSTM. 

\begin{table}[hbtp]
\begin{center}
\resizebox{\linewidth}{!}{
\begin{tabular}{|c|c|c|c|}
 \hline

        Methods & F1@{10,25,50} & mAP@mid & Accuracy \\	
  \hline
	MSN Det \cite{merlshopping} & 46.4, 42.6, 25.6 &\textbf{81.9}  &64.6  \\	  \hline
	MSN Seg \cite{merlshopping} & 80.0, 78.3, 65.4& 69.8 & 76.3  \\	  \hline
	Dilated TCN \cite{leaCVPR} & 79.9, 78.0, 67.5& 75.6 &  76.4 \\	  \hline

	ED- TCN \cite{leaCVPR}& \textbf{86.7}, \textbf{85.1}, \textbf{72.9}& 74.4 & 79.0 \\	  \hline
	Proposed HA\_Net & 80.9, 78.1, 71.2 & 76.7 & \textbf{79.3}  \\	  \hline
			
\end{tabular} }
\end{center}
\vspace{-4mm}
\caption{Action segmentation results for MERL Shopping dataset \cite{merlshopping}. Best values are in bold and the second best values are underlined.}
\label{tab:tab_1}
\end{table}

\begin{table}[hbtp]
\begin{center}
\resizebox{\linewidth}{!}{
\begin{tabular}{|c|c|c|c|}
 \hline

        Methods & F1@{10,25,50} & edit & Accuracy \\	
  \hline
	Spatial CNN \cite{lea2016}&32.3, 27.1, 18.9 & 24.8 & 54.9  \\	  \hline
	 IDT+LM \cite{IDT_LM} &44.4, 38.9, 27.8 & 45.8& 48.7  \\	  \hline
	Dilated TCN \cite{leaCVPR}  & 52.2, 47.6, 37.4 & 43.1 & 59.3 \\	  \hline
	
	ST-CNN \cite{lea2016} &55.9, 49.6, 37.1 & 45.9 & 59.4  \\	  \hline
	Bi-LSTM \cite{leaCVPR}& 62.6, 58.3, 47.0 &55.6 &55.7  \\	  \hline
	ED-TCN \cite{leaCVPR} & 68.0, 63.9, 52.6&59.8 &64.7   \\	  \hline
	TricorNet \cite{ding2017tricornet}& \textbf{70.1}, 67.2, 56.6 & \textbf{62.8}& 67.5 \\	  \hline
	Proposed HA\_Net & 68.2, \textbf{67.3}, \textbf{56.8} & 61.8 & \textbf{67.8} \\ \hline
			
\end{tabular} }
\end{center}
\vspace{-4mm}
\caption{Action segmentation results for 50 salads dataset \cite{stein2013combining}. Best values are in bold and the second best values are underlined. }
\label{tab:tab_2}
\end{table}

\begin{table}[hbtp]
\begin{center}
\resizebox{\linewidth}{!}{
\begin{tabular}{|c|c|c|c|}
 \hline

        Methods & F1@{10,25,50} & edit \\	
  \hline
  
  	EgoNet+TDD \cite{singh2016Egocentric}& NA & 64.4    \\	  \hline                            
	Spatial CNN \cite{lea2016} & 41.8, 36.0, 25.1  &  54.1      \\	  \hline                         
	ST-CNN \cite{lea2016}  & 58.7, 54.4, 41.9& 60.6 \\	  \hline
	Dilated TCN \cite{leaCVPR} & 58.8, 52.2, 42.2 &  58.3    \\	  \hline
	Bi-LSTM \cite{leaCVPR}& 66.5, 59.0, 43.6&  58.3  \\	  \hline
	ED- TCN \cite{leaCVPR}& 72.2, 69.3, 56.0&  64.0 \\	  \hline
	TricorNet \cite{ding2017tricornet}& \textbf{76.0}, \textbf{71.1}, 59.2 &64.8 \\	  \hline
	proposed HA\_Net &73.6, 68.0, \textbf{60.1}  & \textbf{64.3} \\	  \hline
  			
\end{tabular} }
\end{center}
\vspace{-4mm}
\caption{Action segmentation results for Georgia Tech Egocentric dataset \cite{li2015delving}. Best values are in bold and the second best values are underlined.}
\label{tab:tab_3}
\end{table}

The proposed method shows similarity to the TricorNet model by the utilisation of the encoder-decoder framework, and by using LSTMs to map the temporal accordance of the inputs. However, we would like to point out that the proposed method contains fewer trainable parameters when compared to TricorNet due to TricorNet?s use of convolution and bidirectional LSTM layers. The proposed method is very time efficient as a result of hierarchically compressing the input information in the encoding process through the proposed hierarchically attention framework (for more details regarding time complexity please refer to Section 4.9).

Despite being light weight it achieves comparative results to TricorNet due to the careful design of the proposed hierarchical attention framework. The encoder of TricorNet utilises 1D convolutions and pooling to hierarchically embed the features from input frames. However, these fail to capture the temporal relationships between consecutive frames, and fail to determine the salient information that should be passed through the hierarchy when generating feature embeddings. Hence, even though TricorNet is a more complex network, it fails to generate superior results. In contrast, using the proposed hierarchical attention framework we force the network to learn temporal relationships at multiple granularities allowing us to capture an informative feature vector with lower complexity.

\subsection{Ablation Experiment}

To further demonstrate the proposed method we constructed a series of ablation models by strategically removing certain components from the proposed approach (HA\_Net) as follows:
\begin{itemize}
\item{HA\_Net-VE}: We remove the segment-level embedding (see Sec. \ref{sec:segment_level}) component, performing only frame-level embedding.
\item{HA\_Net - (VE, SE)}: From the previous model we removed frame-level attention mechanism (i.e Eq. \ref{eq:2_3} to \ref{eq:2_5}). Therefore the model is a naive LSTM model, which uses only the proposed encoder to simply mapp frame-level embeddings to action classes without using any attention.
\end{itemize}
We evaluated these models against the test set of MERL Shopping dataset \citep{merlshopping} and the evaluation results are presented in Table \ref{tab:tab_4}. When observing the results presented, we observe poor performance from the HA\_Net- (VE, SE) indicating the deficiencies of naive temporal modelling with the recurrent architectures. We improve upon these results in HA\_Net-VE model with the aid of frame level attention through Eq. \ref{eq:2_3} to \ref{eq:2_5}. We would like to compare results of this model with the results presented in Table \ref{tab:tab_1}. It is clear that even frame level attention is sufficient to overcome the shortcomings of the naive RNN models such as MSN Det and MSN Seg. However, with the addition of segment level attention the proposed method maps the temporal relationships at multiple temporal scales, allowing us to overcome the limitations with the TCN architectures such as ED-TCN and TricorNet and outperform all the considered baselines. 

\begin{table}[hbtp]
\begin{center}
\resizebox{\linewidth}{!}{
\begin{tabular}{|c|c|c|c|}
 \hline

        Model & F1@{10,25,50} & mAP@mid & Accuracy \\	
  \hline
  	HA\_Net - (VE, SE) & 72.8, 72.0, 68.4 & 66.5& 71.3 \\ \hline
	HA\_Net - VE& 73.5, 72.8,70.0 & 73.8 & 77.6  \\	  \hline
	HA\_Net & \textbf{80.9, 78.1, 71.2} & \textbf{76.7} & \textbf{79.3}  \\	  \hline
			
\end{tabular} }
\end{center}
\vspace{-6mm}
\caption{HA\_Net  Vs (HA\_Net)-VE (Network without video level embedding) }
\label{tab:tab_4}
\end{table}

\subsection{Hyperparameter evaluation}

\begin{figure*}[htbp]
 \centering
 \begin{subfigure}{.3\textwidth}
       \centering
         \includegraphics[width=.95\linewidth]{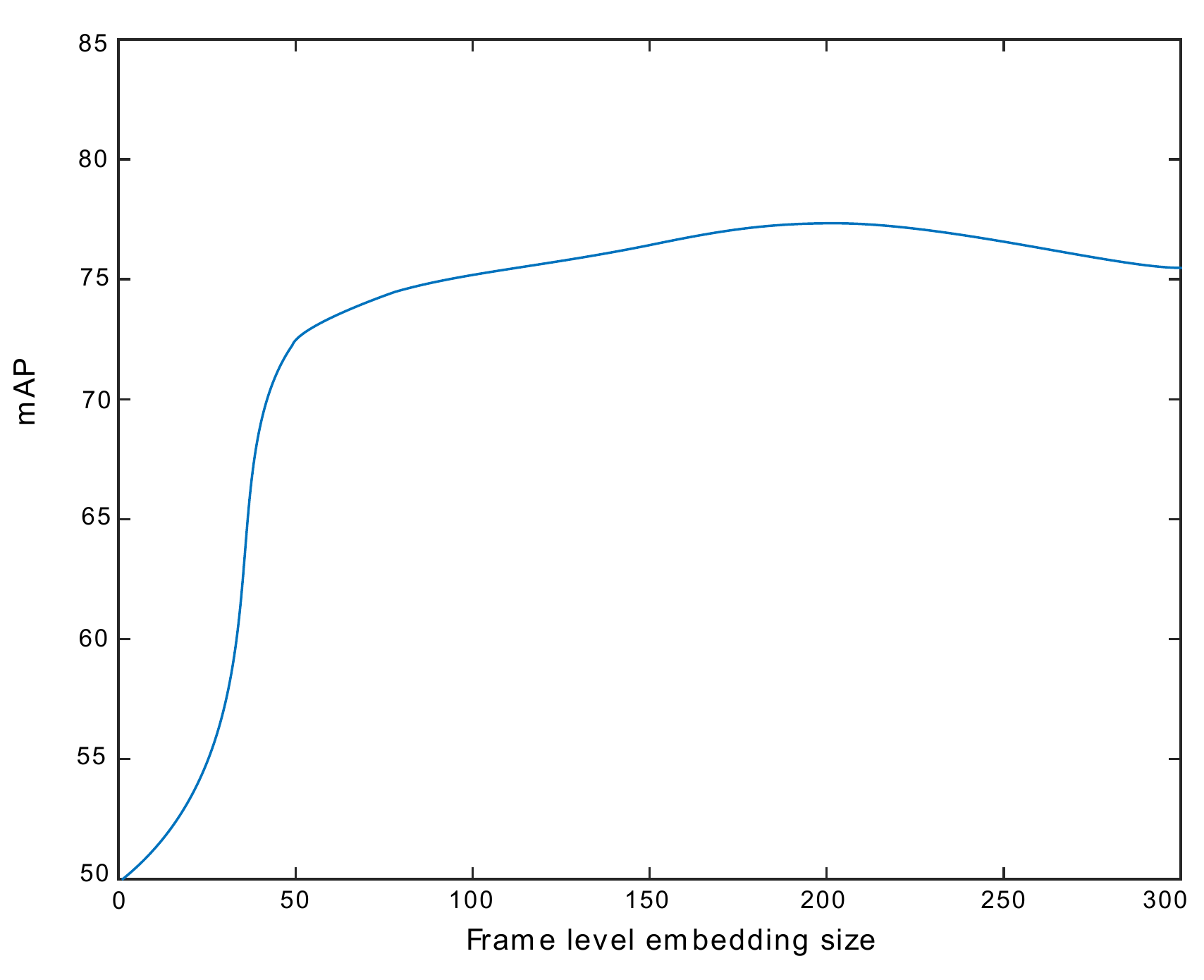} %
          \caption{Embedding dimension ($L$) Vs mAP}
        \end{subfigure}
 \centering
  \begin{subfigure}{.3\textwidth}
       \centering
        \includegraphics[width=.95\linewidth]{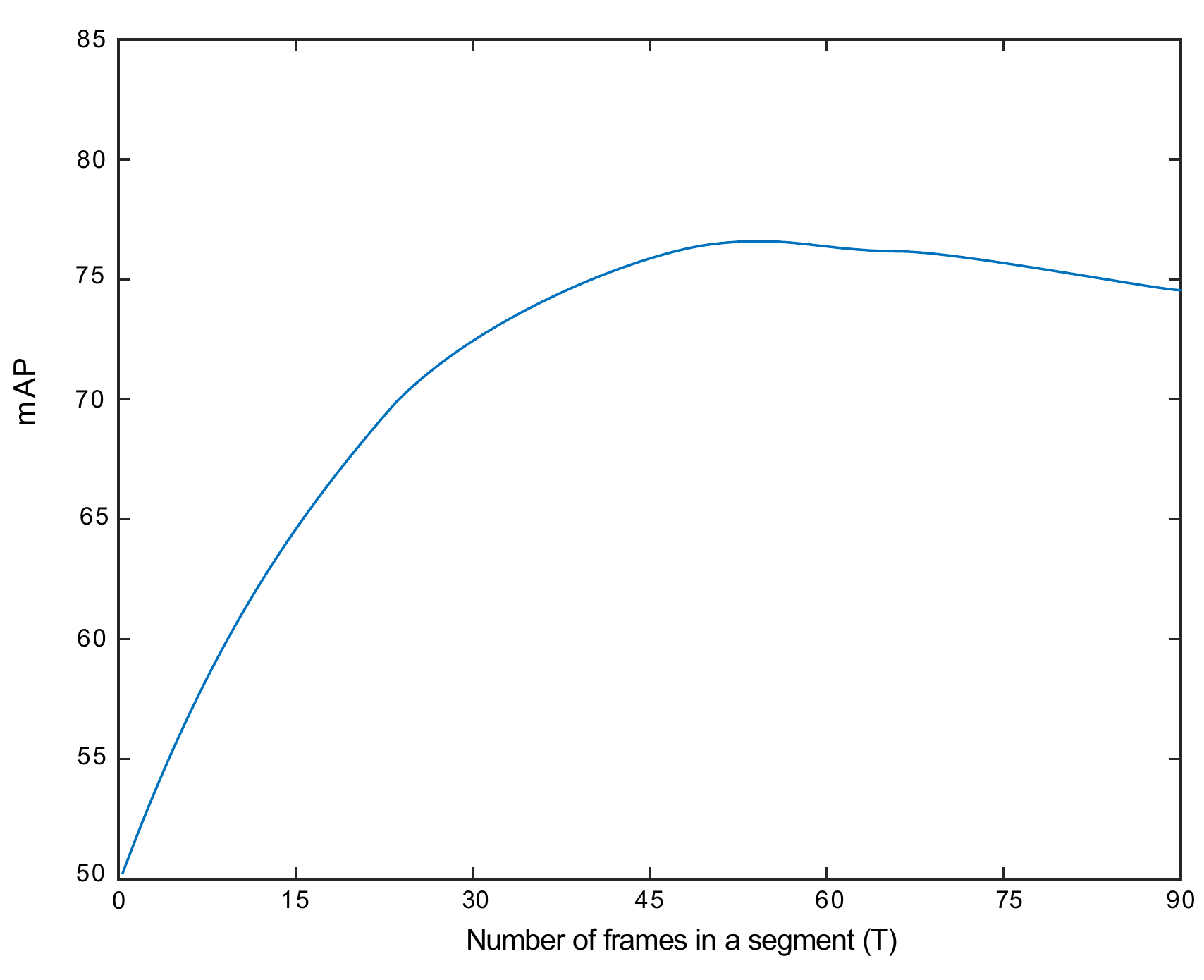} %
        \caption{Number of frames in a segment ($T$) Vs mAP}
    \end{subfigure}
   \centering
      \begin{subfigure}{.3\textwidth}
       \centering
        \includegraphics[width=.95\linewidth]{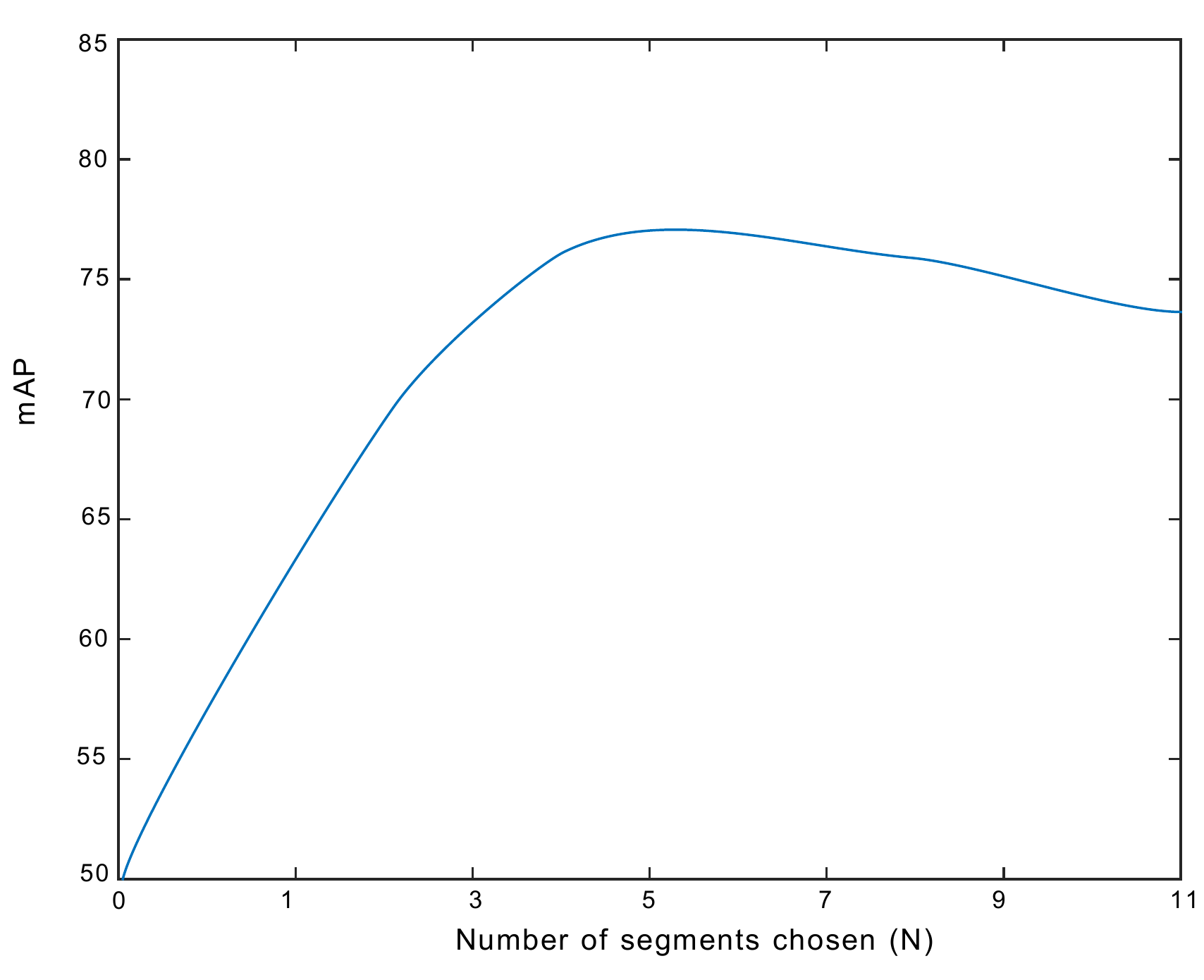} %
         \caption{Number of segments chosen ($N$) Vs mAP}
    \end{subfigure}

    \caption{Evaluation of hyper-parameters: We use the the validation set of the MERL Shopping dataset and tune each parameter  (i.e $L$, $T$ and $N$) while holding the others constant. As $L=200$, $T$=50 and $N$=5 produce best results we utilise these values for evaluating the model for all three datasets.}
        \label{fig:embedding}
\end{figure*}

The frame level, segment level and video level embedding dimensions are of equal size and this embedding dimension $L$ is evaluated experimentally, and is shown in Fig. \ref{fig:embedding} (a). As $L=200$ produces the highest mAP we utilise this as the embeddings size for all the experiments. With a similar experiment we evaluate the number of frames within a segment, $T$; and number of segments $N$, and these results are shown in Fig. \ref{fig:embedding} (b) and (c), respectively. As such $T$ is set to be 50 and $N$ is chosen to be $5$. In these experiments we use the the validation set of the MERL Shopping dataset.  We tune each parameter while holding the others constant.   

As shown in Fig. \ref{fig:embedding} (b), when the number of frames that are considered is less the accuracies are lower, due to the fact that few frames provides less temporal cues for the action segmentation. Hence, with the number of frames (T), the accuracy gradually increases until T=50. Beyond that point the accuracy tends to decrease slowly.  This is mainly due to the reason that when mapping longer sequences into a finite size frame level vector through an LSTM can cause information loss. However when considering short segments of the sequence,  utilising the proposed attention mechanism, we can map the fine-grained details in to the frame-level embedding vector.

In Fig. \ref{fig:embedding} (c), when the number of segments is set lower, the segment embedding contains less information which makes the decoding process more difficult. Also when the number of segments is higher, it tends to confuse the model with too much information. When evaluating, we found that the optimal value for N=5, which provides satisfactory information for the segmentation process. These values of T=50 and N=5 have been utilised when evaluating the model for all three datasets.

\subsection{ Qualitative results}
\begin{figure*}[htbp]
 \centering
 \begin{subfigure}{.46\textwidth}
       \centering
         \includegraphics[width=.99\linewidth]{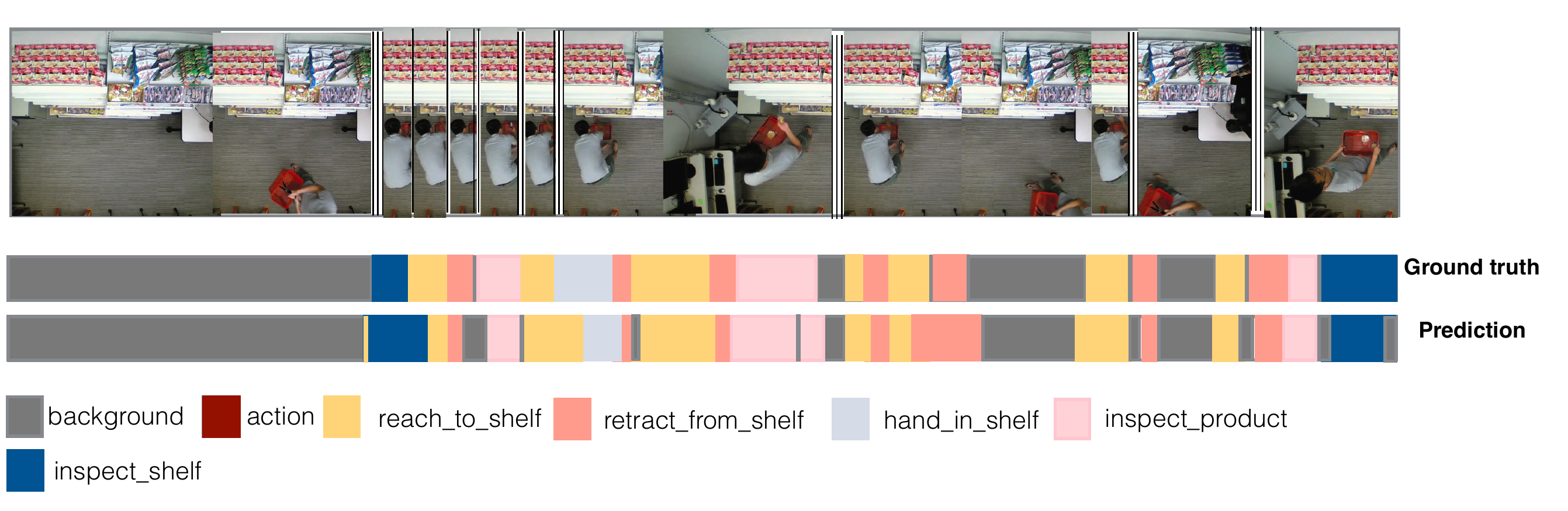} %
          \caption{}
        \end{subfigure}
 \centering
  \begin{subfigure}{.46\textwidth}
       \centering
        \includegraphics[width=.99\linewidth]{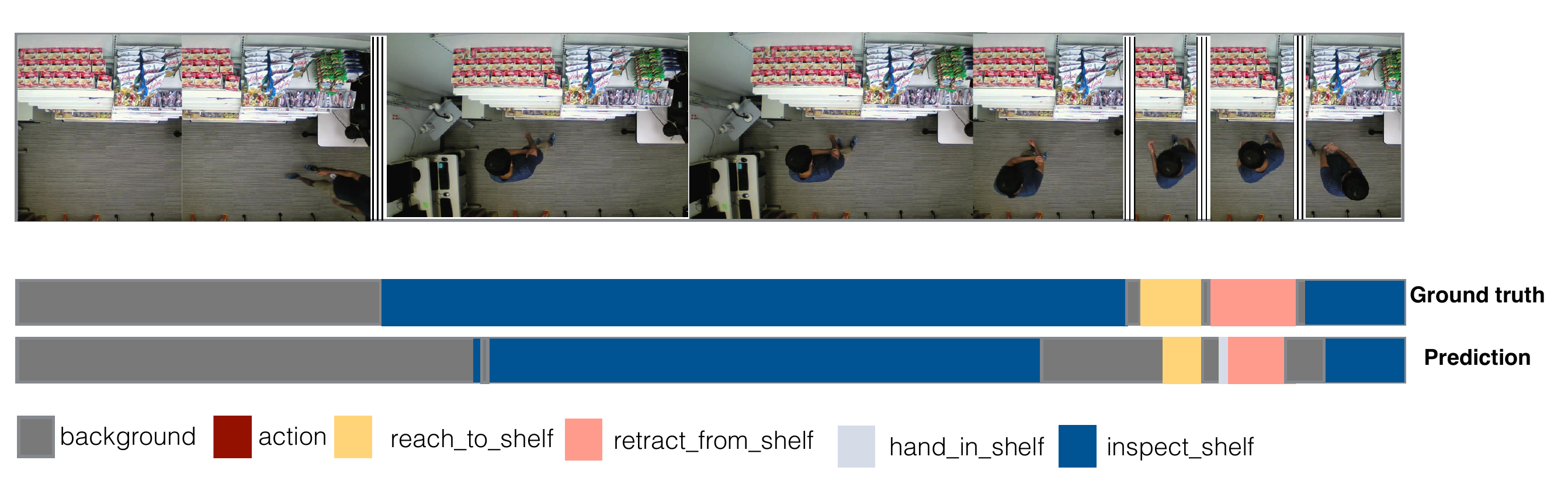} %
        \caption{}
    \end{subfigure}
\vspace{-4.5mm}
    \caption{Action segmentation results for MERL Shopping \cite{merlshopping} dataset}
        \label{fig:pred3}
\end{figure*}

\begin{figure*}[htbp]
 \centering
 \begin{subfigure}{.46\textwidth}
       \centering
         \includegraphics[width=.99\linewidth]{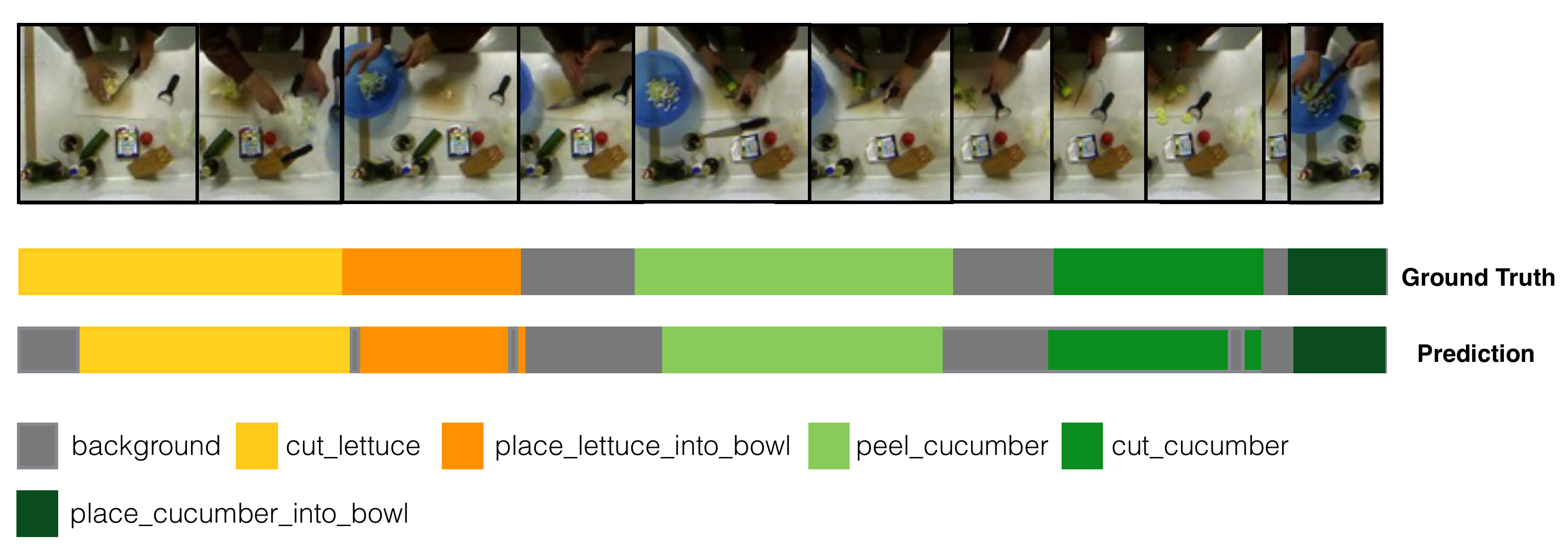} %
          \caption{}
        \end{subfigure}
 \centering
  \begin{subfigure}{.46\textwidth}
       \centering
        \includegraphics[width=.99\linewidth]{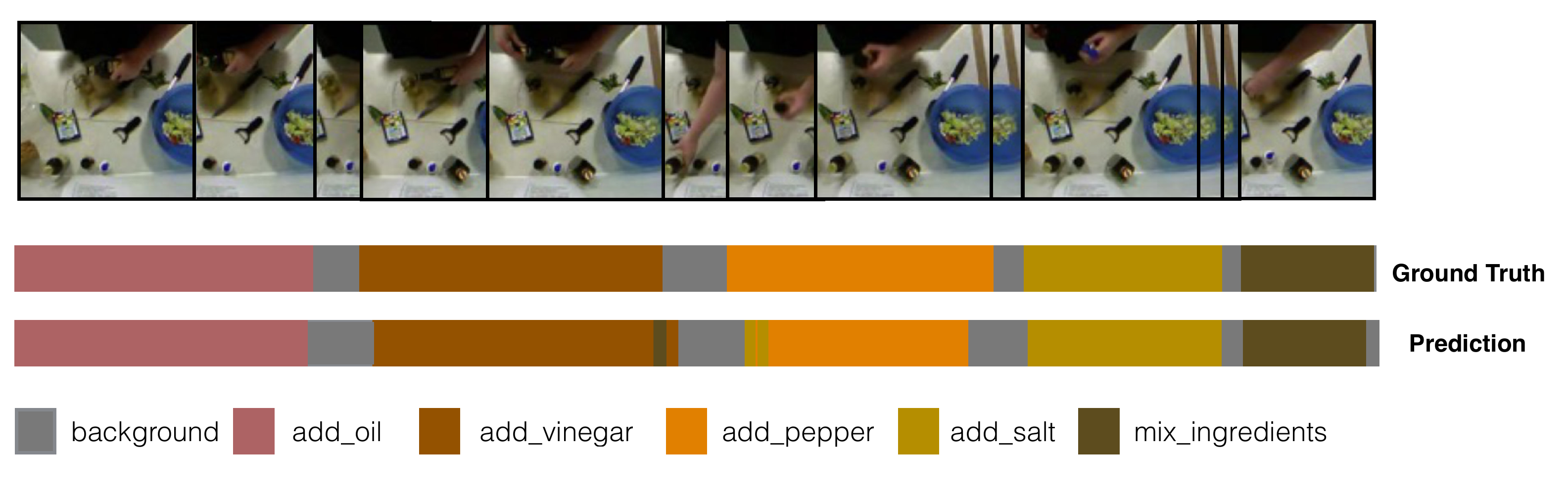} %
        \caption{}
    \end{subfigure}
\vspace{-4.5mm}
    \caption{Action segmentation results for 50 salads \cite{stein2013combining} dataset}
        \label{fig:pred4}
\end{figure*}

\begin{figure*}[htbp]
 \centering
 \begin{subfigure}{.46\textwidth}
       \centering
         \includegraphics[width=.99\linewidth]{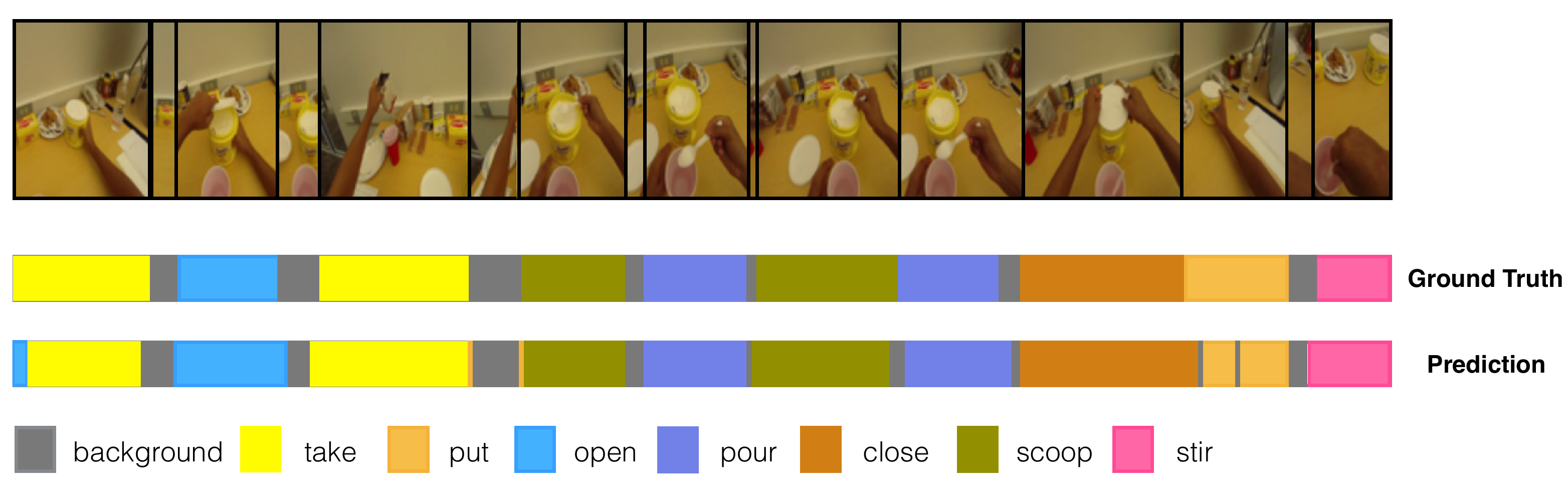} %
          \caption{}
        \end{subfigure}
 \centering
  \begin{subfigure}{.46\textwidth}
       \centering
        \includegraphics[width=.99\linewidth]{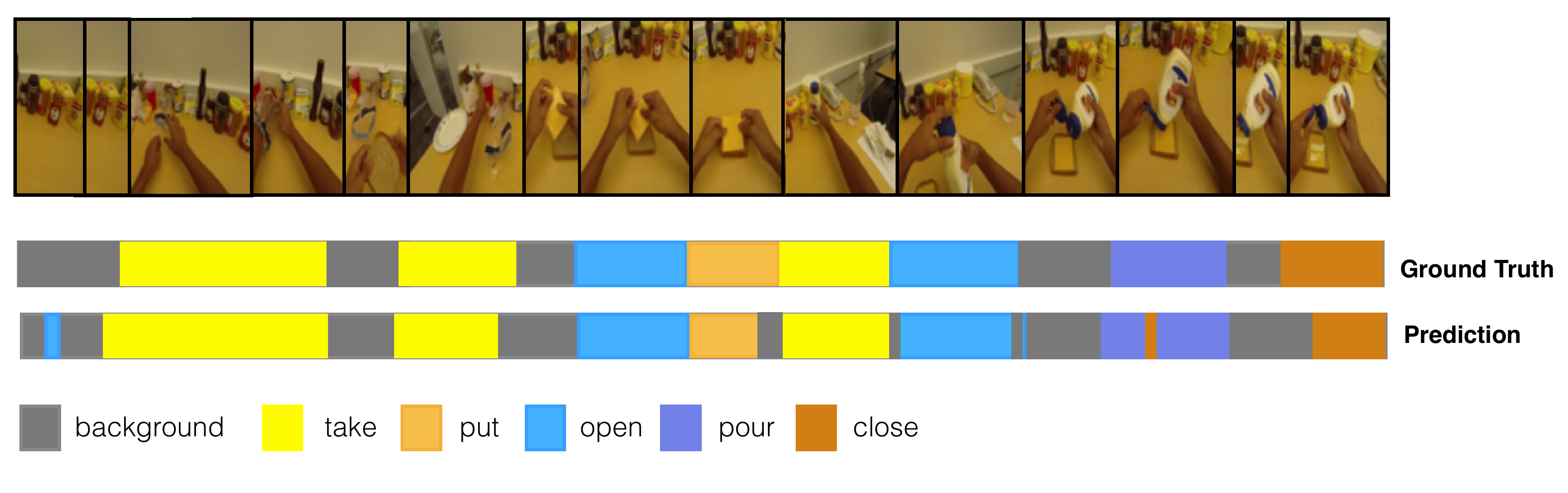} %
        \caption{}
    \end{subfigure}
\vspace{-4.5mm}
    \caption{Action segmentation results for Georgia Tech Egocentric \cite{li2015delving} dataset}
        \label{fig:pred5}
\end{figure*}

Qualitative results of the proposed model on MERL Shopping \citep{merlshopping}, 50 salads \citep{stein2013combining} and Georgia Tech Egocentric \citep{li2015delving} datasets are presented in Figures \ref{fig:pred3}, \ref{fig:pred4} and \ref{fig:pred5}, respectively. We observe slight confusion between action classes, typically at the event boundaries where it is quite difficult to determine the action transitions. However in general we observe that all the action classes are being detected and false detections only last for short periods of time.

\subsection{Time efficiency}
 The proposed approach has only 13.5M parameters. We ran the test set of the MERL shopping dataset on single core of an Intel E5-2680 2.50GHz CPU and the model generates 1000 predictions, with $50 \times 5 = 250$ frames in each prediction, in 21.2 seconds.

\section{Conclusion}
\label{conclusion}
We have introduced a novel hierarchical attention framework for segmenting human actions in videos. The proposed framework spans it's attention across multiple temporal scales, embedding the context information as a vector representation of the video. The decoder model hierarchically decodes this embedded information, effectively propagating the salient aspects to the classification module. Our empirical evaluations on MERL Shopping, 50 salads, and Georgia Tech Egocentric datasets clearly demonstrates the effectiveness of the proposed method in different application domains including intelligent surveillance and home monitoring. The ablation experiments suggest the importance of the bottom up computational process, through frame level to video level compression of the video embeddings, allowing the model to explicitly learn the long-term temporal dependencies. 

\bibliographystyle{model2-names}
\bibliography{paper2}

\begin{thebibliography}{44}
\expandafter\ifx\csname natexlab\endcsname\relax\def\natexlab#1{#1}\fi
\providecommand{\url}[1]{\texttt{#1}}
\providecommand{\href}[2]{#2}
\providecommand{\path}[1]{#1}
\providecommand{\DOIprefix}{doi:}
\providecommand{\ArXivprefix}{arXiv:}
\providecommand{\URLprefix}{URL: }
\providecommand{\Pubmedprefix}{pmid:}
\providecommand{\doi}[1]{\href{http://dx.doi.org/#1}{\path{#1}}}
\providecommand{\Pubmed}[1]{\href{pmid:#1}{\path{#1}}}
\providecommand{\bibinfo}[2]{#2}
\ifx\xfnm\relax \def\xfnm[#1]{\unskip,\space#1}\fi
\bibitem[{Bergstra et~al.(2010)Bergstra, Breuleux, Bastien, Lamblin, Pascanu,
  Desjardins, Turian, Warde-Farley and Bengio}]{bergstra2010theano}
\bibinfo{author}{Bergstra, J.}, \bibinfo{author}{Breuleux, O.},
  \bibinfo{author}{Bastien, F.}, \bibinfo{author}{Lamblin, P.},
  \bibinfo{author}{Pascanu, R.}, \bibinfo{author}{Desjardins, G.},
  \bibinfo{author}{Turian, J.}, \bibinfo{author}{Warde-Farley, D.},
  \bibinfo{author}{Bengio, Y.}, \bibinfo{year}{2010}.
\newblock \bibinfo{title}{Theano: A cpu and gpu math compiler in python}, in:
  \bibinfo{booktitle}{Proc. 9th Python in Science Conf}.
\bibitem[{Chollet et~al.(2015)}]{chollet2015keras}
\bibinfo{author}{Chollet, F.}, et~al., \bibinfo{year}{2015}.
\newblock \bibinfo{title}{Keras}.
\bibitem[{Delaitre et~al.(2010)Delaitre, Laptev and Sivic}]{imageBased1}
\bibinfo{author}{Delaitre, V.}, \bibinfo{author}{Laptev, I.},
  \bibinfo{author}{Sivic, J.}, \bibinfo{year}{2010}.
\newblock \bibinfo{title}{Recognizing human actions in still images: a study of
  bag-of-features and part-based representations}.
\newblock \bibinfo{note}{Updated version, available at
  http://www.di.ens.fr/willow/research/stillactions/}.
\bibitem[{Deng et~al.(2016a)Deng, Bao, Kong, Ren and Dai}]{deng2016deep}
\bibinfo{author}{Deng, Y.}, \bibinfo{author}{Bao, F.}, \bibinfo{author}{Kong,
  Y.}, \bibinfo{author}{Ren, Z.}, \bibinfo{author}{Dai, Q.},
  \bibinfo{year}{2016}a.
\newblock \bibinfo{title}{Deep direct reinforcement learning for financial
  signal representation and trading}.
\newblock \bibinfo{journal}{IEEE transactions on neural networks and learning
  systems} \bibinfo{volume}{28}, \bibinfo{pages}{653--664}.
\bibitem[{Deng et~al.(2013)Deng, Dai, Liu, Zhang and Hu}]{deng2013low}
\bibinfo{author}{Deng, Y.}, \bibinfo{author}{Dai, Q.}, \bibinfo{author}{Liu,
  R.}, \bibinfo{author}{Zhang, Z.}, \bibinfo{author}{Hu, S.},
  \bibinfo{year}{2013}.
\newblock \bibinfo{title}{Low-rank structure learning via nonconvex heuristic
  recovery}.
\newblock \bibinfo{journal}{IEEE transactions on neural networks and learning
  systems} \bibinfo{volume}{24}, \bibinfo{pages}{383--396}.
\bibitem[{Deng et~al.(2016b)Deng, Ren, Kong, Bao and
  Dai}]{deng2016hierarchical}
\bibinfo{author}{Deng, Y.}, \bibinfo{author}{Ren, Z.}, \bibinfo{author}{Kong,
  Y.}, \bibinfo{author}{Bao, F.}, \bibinfo{author}{Dai, Q.},
  \bibinfo{year}{2016}b.
\newblock \bibinfo{title}{A hierarchical fused fuzzy deep neural network for
  data classification}.
\newblock \bibinfo{journal}{IEEE Transactions on Fuzzy Systems}
  \bibinfo{volume}{25}, \bibinfo{pages}{1006--1012}.
\bibitem[{Ding and Xu(2017)}]{ding2017tricornet}
\bibinfo{author}{Ding, L.}, \bibinfo{author}{Xu, C.}, \bibinfo{year}{2017}.
\newblock \bibinfo{title}{Tricornet: A hybrid temporal convolutional and
  recurrent network for video action segmentation}.
\newblock \bibinfo{journal}{arXiv preprint arXiv:1705.07818} .
\bibitem[{Fernando et~al.(2015)Fernando, Gavves, Oramas, Ghodrati and
  Tuytelaars}]{videoDarwin}
\bibinfo{author}{Fernando, B.}, \bibinfo{author}{Gavves, E.},
  \bibinfo{author}{Oramas, J.M.}, \bibinfo{author}{Ghodrati, A.},
  \bibinfo{author}{Tuytelaars, T.}, \bibinfo{year}{2015}.
\newblock \bibinfo{title}{Modeling video evolution for action recognition}, in:
  \bibinfo{booktitle}{Proceedings of the IEEE Conference on Computer Vision and
  Pattern Recognition}, pp. \bibinfo{pages}{5378--5387}.
\bibitem[{Fernando et~al.(2018)Fernando, Denman, McFadyen, Sridharan and
  Fookes}]{fernando2018tree}
\bibinfo{author}{Fernando, T.}, \bibinfo{author}{Denman, S.},
  \bibinfo{author}{McFadyen, A.}, \bibinfo{author}{Sridharan, S.},
  \bibinfo{author}{Fookes, C.}, \bibinfo{year}{2018}.
\newblock \bibinfo{title}{Tree memory networks for modelling long-term temporal
  dependencies}.
\newblock \bibinfo{journal}{Neurocomputing} \bibinfo{volume}{304},
  \bibinfo{pages}{64--81}.
\bibitem[{Gammulle et~al.(2017)Gammulle, Denman, Sridharan and
  Fookes}]{gammulle2017two}
\bibinfo{author}{Gammulle, H.}, \bibinfo{author}{Denman, S.},
  \bibinfo{author}{Sridharan, S.}, \bibinfo{author}{Fookes, C.},
  \bibinfo{year}{2017}.
\newblock \bibinfo{title}{Two stream lstm: A deep fusion framework for human
  action recognition}, in: \bibinfo{booktitle}{Applications of Computer Vision
  (WACV), 2017 IEEE Winter Conference on}, \bibinfo{organization}{IEEE}. pp.
  \bibinfo{pages}{177--186}.
\bibitem[{Gammulle et~al.(2019a)Gammulle, Denman, Sridharan and
  Fookes}]{gammulle2019forecasting}
\bibinfo{author}{Gammulle, H.}, \bibinfo{author}{Denman, S.},
  \bibinfo{author}{Sridharan, S.}, \bibinfo{author}{Fookes, C.},
  \bibinfo{year}{2019}a.
\newblock \bibinfo{title}{Forecasting future action sequences with neural
  memory networks}.
\newblock \bibinfo{journal}{arXiv preprint arXiv:1909.09278} .
\bibitem[{Gammulle et~al.(2019b)Gammulle, Denman, Sridharan and
  Fookes}]{gammulle2019predicting}
\bibinfo{author}{Gammulle, H.}, \bibinfo{author}{Denman, S.},
  \bibinfo{author}{Sridharan, S.}, \bibinfo{author}{Fookes, C.},
  \bibinfo{year}{2019}b.
\newblock \bibinfo{title}{Predicting the future: A jointly learnt model for
  action anticipation}, in: \bibinfo{booktitle}{Proceedings of the IEEE
  International Conference on Computer Vision}, pp.
  \bibinfo{pages}{5562--5571}.
\bibitem[{Gammulle et~al.(2020)Gammulle, Denman, Sridharan and
  Fookes}]{gammulle2020fine}
\bibinfo{author}{Gammulle, H.}, \bibinfo{author}{Denman, S.},
  \bibinfo{author}{Sridharan, S.}, \bibinfo{author}{Fookes, C.},
  \bibinfo{year}{2020}.
\newblock \bibinfo{title}{Fine-grained action segmentation using the
  semi-supervised action gan}.
\newblock \bibinfo{journal}{Pattern Recognition} \bibinfo{volume}{98},
  \bibinfo{pages}{107039}.
\bibitem[{He et~al.(2016)He, Zhang, Ren and Sun}]{resnet}
\bibinfo{author}{He, K.}, \bibinfo{author}{Zhang, X.}, \bibinfo{author}{Ren,
  S.}, \bibinfo{author}{Sun, J.}, \bibinfo{year}{2016}.
\newblock \bibinfo{title}{Deep residual learning for image recognition}, in:
  \bibinfo{booktitle}{Proceedings of the IEEE Conference on Computer Vision and
  Pattern Recognition}, pp. \bibinfo{pages}{770--778}.
\bibitem[{Kingma and Ba(2014)}]{adam}
\bibinfo{author}{Kingma, D.}, \bibinfo{author}{Ba, J.}, \bibinfo{year}{2014}.
\newblock \bibinfo{title}{Adam: A method for stochastic optimization}.
\newblock \bibinfo{journal}{arXiv preprint arXiv:1412.6980} .
\bibitem[{Kumar et~al.(2016)Kumar, Irsoy, Ondruska, Iyyer, Bradbury, Gulrajani,
  Zhong, Paulus and Socher}]{kumar2016ask}
\bibinfo{author}{Kumar, A.}, \bibinfo{author}{Irsoy, O.},
  \bibinfo{author}{Ondruska, P.}, \bibinfo{author}{Iyyer, M.},
  \bibinfo{author}{Bradbury, J.}, \bibinfo{author}{Gulrajani, I.},
  \bibinfo{author}{Zhong, V.}, \bibinfo{author}{Paulus, R.},
  \bibinfo{author}{Socher, R.}, \bibinfo{year}{2016}.
\newblock \bibinfo{title}{Ask me anything: Dynamic memory networks for natural
  language processing}, in: \bibinfo{booktitle}{International Conference on
  Machine Learning}, pp. \bibinfo{pages}{1378--1387}.
\bibitem[{de~La~Gorce et~al.(2008)de~La~Gorce, Paragios and Fleet}]{de2008}
\bibinfo{author}{de~La~Gorce, M.}, \bibinfo{author}{Paragios, N.},
  \bibinfo{author}{Fleet, D.J.}, \bibinfo{year}{2008}.
\newblock \bibinfo{title}{Model-based hand tracking with texture, shading and
  self-occlusions}, in: \bibinfo{booktitle}{Computer Vision and Pattern
  Recognition, 2008. CVPR 2008. IEEE Conference On},
  \bibinfo{organization}{IEEE}. pp. \bibinfo{pages}{1--8}.
\bibitem[{Lea et~al.(2017)Lea, Flynn, Vidal, Reiter and Hager}]{leaCVPR}
\bibinfo{author}{Lea, C.}, \bibinfo{author}{Flynn, M.D.},
  \bibinfo{author}{Vidal, R.}, \bibinfo{author}{Reiter, A.},
  \bibinfo{author}{Hager, G.D.}, \bibinfo{year}{2017}.
\newblock \bibinfo{title}{Temporal convolutional networks for action
  segmentation and detection}, in: \bibinfo{booktitle}{2017 IEEE Conference on
  Computer Vision and Pattern Recognition (CVPR)}.
\bibitem[{Lea et~al.(2016a)Lea, Reiter, Vidal and Hager}]{lea2016}
\bibinfo{author}{Lea, C.}, \bibinfo{author}{Reiter, A.},
  \bibinfo{author}{Vidal, R.}, \bibinfo{author}{Hager, G.D.},
  \bibinfo{year}{2016}a.
\newblock \bibinfo{title}{Segmental spatiotemporal cnns for fine-grained action
  segmentation}, in: \bibinfo{booktitle}{European Conference on Computer
  Vision}, \bibinfo{organization}{Springer}. pp. \bibinfo{pages}{36--52}.
\bibitem[{Lea et~al.(2016b)Lea, Vidal and Hager}]{lea2016learning}
\bibinfo{author}{Lea, C.}, \bibinfo{author}{Vidal, R.}, \bibinfo{author}{Hager,
  G.D.}, \bibinfo{year}{2016}b.
\newblock \bibinfo{title}{Learning convolutional action primitives for
  fine-grained action recognition}, in: \bibinfo{booktitle}{Robotics and
  Automation (ICRA), 2016 IEEE International Conference on},
  \bibinfo{organization}{IEEE}. pp. \bibinfo{pages}{1642--1649}.
\bibitem[{Lea et~al.(2016c)Lea, Vidal, Reiter and Hager}]{lea2016temporal}
\bibinfo{author}{Lea, C.}, \bibinfo{author}{Vidal, R.},
  \bibinfo{author}{Reiter, A.}, \bibinfo{author}{Hager, G.D.},
  \bibinfo{year}{2016}c.
\newblock \bibinfo{title}{Temporal convolutional networks: A unified approach
  to action segmentation}, in: \bibinfo{booktitle}{European Conference on
  Computer Vision}, \bibinfo{organization}{Springer}. pp.
  \bibinfo{pages}{47--54}.
\bibitem[{Li et~al.(2015a)Li, Luong and Jurafsky}]{hierLSTM}
\bibinfo{author}{Li, J.}, \bibinfo{author}{Luong, M.},
  \bibinfo{author}{Jurafsky, D.}, \bibinfo{year}{2015}a.
\newblock \bibinfo{title}{A hierarchical neural autoencoder for paragraphs and
  documents}.
\newblock \bibinfo{journal}{CoRR} \bibinfo{volume}{abs/1506.01057}.
\bibitem[{Li et~al.(2015b)Li, Ye and Rehg}]{li2015delving}
\bibinfo{author}{Li, Y.}, \bibinfo{author}{Ye, Z.}, \bibinfo{author}{Rehg,
  J.M.}, \bibinfo{year}{2015}b.
\newblock \bibinfo{title}{Delving into egocentric actions}, in:
  \bibinfo{booktitle}{Proceedings of the IEEE Conference on Computer Vision and
  Pattern Recognition}, pp. \bibinfo{pages}{287--295}.
\bibitem[{Ni et~al.(2014)Ni, Paramathayalan and Moulin}]{Ni2014}
\bibinfo{author}{Ni, B.}, \bibinfo{author}{Paramathayalan, V.R.},
  \bibinfo{author}{Moulin, P.}, \bibinfo{year}{2014}.
\newblock \bibinfo{title}{Multiple granularity analysis for fine-grained action
  detection}, in: \bibinfo{booktitle}{The IEEE Conference on Computer Vision
  and Pattern Recognition (CVPR)}.
\bibitem[{Richard and Gall(2016)}]{IDT_LM}
\bibinfo{author}{Richard, A.}, \bibinfo{author}{Gall, J.},
  \bibinfo{year}{2016}.
\newblock \bibinfo{title}{Temporal action detection using a statistical
  language model}, in: \bibinfo{booktitle}{Proceedings of the IEEE Conference
  on Computer Vision and Pattern Recognition}, pp. \bibinfo{pages}{3131--3140}.
\bibitem[{Rohrbach et~al.(2012)Rohrbach, Amin, Andriluka and
  Schiele}]{cooking1}
\bibinfo{author}{Rohrbach, M.}, \bibinfo{author}{Amin, S.},
  \bibinfo{author}{Andriluka, M.}, \bibinfo{author}{Schiele, B.},
  \bibinfo{year}{2012}.
\newblock \bibinfo{title}{A database for fine grained activity detection of
  cooking activities}, in: \bibinfo{booktitle}{2012 IEEE Conference on Computer
  Vision and Pattern Recognition (CVPR)}, \bibinfo{address}{Providence, United
  States}.
\bibitem[{Rohrbach et~al.(2016)Rohrbach, Rohrbach, Regneri, Amin, Andriluka,
  Pinkal and Schiele}]{rohrbach2016}
\bibinfo{author}{Rohrbach, M.}, \bibinfo{author}{Rohrbach, A.},
  \bibinfo{author}{Regneri, M.}, \bibinfo{author}{Amin, S.},
  \bibinfo{author}{Andriluka, M.}, \bibinfo{author}{Pinkal, M.},
  \bibinfo{author}{Schiele, B.}, \bibinfo{year}{2016}.
\newblock \bibinfo{title}{Recognizing fine-grained and composite activities
  using hand-centric features and script data}.
\newblock \bibinfo{journal}{International Journal of Computer Vision}
  \bibinfo{volume}{119}, \bibinfo{pages}{346--373}.
\bibitem[{Russakovsky et~al.(2015)Russakovsky, Deng, Su, Krause, Satheesh, Ma,
  Huang, Karpathy, Khosla, Bernstein, Berg and Fei-Fei}]{imageNet}
\bibinfo{author}{Russakovsky, O.}, \bibinfo{author}{Deng, J.},
  \bibinfo{author}{Su, H.}, \bibinfo{author}{Krause, J.},
  \bibinfo{author}{Satheesh, S.}, \bibinfo{author}{Ma, S.},
  \bibinfo{author}{Huang, Z.}, \bibinfo{author}{Karpathy, A.},
  \bibinfo{author}{Khosla, A.}, \bibinfo{author}{Bernstein, M.},
  \bibinfo{author}{Berg, A.C.}, \bibinfo{author}{Fei-Fei, L.},
  \bibinfo{year}{2015}.
\newblock \bibinfo{title}{{ImageNet Large Scale Visual Recognition Challenge}}.
\newblock \bibinfo{journal}{International Journal of Computer Vision (IJCV)}
  \bibinfo{volume}{115}, \bibinfo{pages}{211--252}.
\bibitem[{Simonyan and Zisserman(2014)}]{twoStream}
\bibinfo{author}{Simonyan, K.}, \bibinfo{author}{Zisserman, A.},
  \bibinfo{year}{2014}.
\newblock \bibinfo{title}{Two-stream convolutional networks for action
  recognition in videos}.
\newblock \bibinfo{journal}{CoRR} \bibinfo{volume}{abs/1406.2199}.
\bibitem[{Singh et~al.(2016a)Singh, Marks, Jones, Tuzel and
  Shao}]{merlshopping}
\bibinfo{author}{Singh, B.}, \bibinfo{author}{Marks, T.K.},
  \bibinfo{author}{Jones, M.}, \bibinfo{author}{Tuzel, O.},
  \bibinfo{author}{Shao, M.}, \bibinfo{year}{2016}a.
\newblock \bibinfo{title}{A multi-stream bi-directional recurrent neural
  network for fine-grained action detection}, in: \bibinfo{booktitle}{The IEEE
  Conference on Computer Vision and Pattern Recognition (CVPR)}.
\bibitem[{Singh et~al.(2016b)Singh, Arora and Jawahar}]{singh2016Egocentric}
\bibinfo{author}{Singh, S.}, \bibinfo{author}{Arora, C.},
  \bibinfo{author}{Jawahar, C.}, \bibinfo{year}{2016}b.
\newblock \bibinfo{title}{First person action recognition using deep learned
  descriptors}, in: \bibinfo{booktitle}{Proceedings of the IEEE Conference on
  Computer Vision and Pattern Recognition}, pp. \bibinfo{pages}{2620--2628}.
\bibitem[{Sordoni et~al.(2015)Sordoni, Bengio, Vahabi, Lioma, Grue~Simonsen and
  Nie}]{sordoni2015hierarchical}
\bibinfo{author}{Sordoni, A.}, \bibinfo{author}{Bengio, Y.},
  \bibinfo{author}{Vahabi, H.}, \bibinfo{author}{Lioma, C.},
  \bibinfo{author}{Grue~Simonsen, J.}, \bibinfo{author}{Nie, J.Y.},
  \bibinfo{year}{2015}.
\newblock \bibinfo{title}{A hierarchical recurrent encoder-decoder for
  generative context-aware query suggestion}, in:
  \bibinfo{booktitle}{Proceedings of the 24th ACM International on Conference
  on Information and Knowledge Management}, \bibinfo{organization}{ACM}. pp.
  \bibinfo{pages}{553--562}.
\bibitem[{Stein and McKenna(2013)}]{stein2013combining}
\bibinfo{author}{Stein, S.}, \bibinfo{author}{McKenna, S.J.},
  \bibinfo{year}{2013}.
\newblock \bibinfo{title}{Combining embedded accelerometers with computer
  vision for recognizing food preparation activities}, in:
  \bibinfo{booktitle}{Proceedings of the 2013 ACM international joint
  conference on Pervasive and ubiquitous computing},
  \bibinfo{organization}{ACM}. pp. \bibinfo{pages}{729--738}.
\bibitem[{Trinh et~al.(2012)Trinh, Fan, Gabbur and Pankanti}]{trinh2012}
\bibinfo{author}{Trinh, H.}, \bibinfo{author}{Fan, Q.},
  \bibinfo{author}{Gabbur, P.}, \bibinfo{author}{Pankanti, S.},
  \bibinfo{year}{2012}.
\newblock \bibinfo{title}{Hand tracking by binary quadratic programming and its
  application to retail activity recognition}, in: \bibinfo{booktitle}{Computer
  Vision and Pattern Recognition (CVPR), 2012 IEEE Conference on},
  \bibinfo{organization}{IEEE}. pp. \bibinfo{pages}{1902--1909}.
\bibitem[{Wu et~al.(2018a)Wu, Wang, Gao and Li}]{wu2018deep}
\bibinfo{author}{Wu, L.}, \bibinfo{author}{Wang, Y.}, \bibinfo{author}{Gao,
  J.}, \bibinfo{author}{Li, X.}, \bibinfo{year}{2018}a.
\newblock \bibinfo{title}{Deep adaptive feature embedding with local sample
  distributions for person re-identification}.
\newblock \bibinfo{journal}{Pattern Recognition} \bibinfo{volume}{73},
  \bibinfo{pages}{275--288}.
\bibitem[{Wu et~al.(2018b)Wu, Wang, Gao and Li}]{wu2018and2}
\bibinfo{author}{Wu, L.}, \bibinfo{author}{Wang, Y.}, \bibinfo{author}{Gao,
  J.}, \bibinfo{author}{Li, X.}, \bibinfo{year}{2018}b.
\newblock \bibinfo{title}{Where-and-when to look: Deep siamese attention
  networks for video-based person re-identification}.
\newblock \bibinfo{journal}{IEEE Transactions on Multimedia}
  \bibinfo{volume}{21}, \bibinfo{pages}{1412--1424}.
\bibitem[{Wu et~al.(2018c)Wu, Wang, Li and Gao}]{wu2018deep2}
\bibinfo{author}{Wu, L.}, \bibinfo{author}{Wang, Y.}, \bibinfo{author}{Li, X.},
  \bibinfo{author}{Gao, J.}, \bibinfo{year}{2018}c.
\newblock \bibinfo{title}{Deep attention-based spatially recursive networks for
  fine-grained visual recognition}.
\newblock \bibinfo{journal}{IEEE transactions on cybernetics}
  \bibinfo{volume}{49}, \bibinfo{pages}{1791--1802}.
\bibitem[{Wu et~al.(2018d)Wu, Wang, Li and Gao}]{wu2018and}
\bibinfo{author}{Wu, L.}, \bibinfo{author}{Wang, Y.}, \bibinfo{author}{Li, X.},
  \bibinfo{author}{Gao, J.}, \bibinfo{year}{2018}d.
\newblock \bibinfo{title}{What-and-where to match: Deep spatially
  multiplicative integration networks for person re-identification}.
\newblock \bibinfo{journal}{Pattern Recognition} \bibinfo{volume}{76},
  \bibinfo{pages}{727--738}.
\bibitem[{Yang et~al.(2017)Yang, Zhou, Fan, Gao, Wu, Ou, Lu, Cheng and
  Latecki}]{yang2017deep}
\bibinfo{author}{Yang, W.}, \bibinfo{author}{Zhou, Q.}, \bibinfo{author}{Fan,
  Y.}, \bibinfo{author}{Gao, G.}, \bibinfo{author}{Wu, S.},
  \bibinfo{author}{Ou, W.}, \bibinfo{author}{Lu, H.}, \bibinfo{author}{Cheng,
  J.}, \bibinfo{author}{Latecki, L.J.}, \bibinfo{year}{2017}.
\newblock \bibinfo{title}{Deep context convolutional neural networks for
  semantic segmentation}, in: \bibinfo{booktitle}{CCF Chinese Conference on
  computer vision}, \bibinfo{organization}{Springer}. pp.
  \bibinfo{pages}{696--704}.
\bibitem[{You et~al.(2014)You, Li, Tao, Ou and Gong}]{you2014local}
\bibinfo{author}{You, X.}, \bibinfo{author}{Li, Q.}, \bibinfo{author}{Tao, D.},
  \bibinfo{author}{Ou, W.}, \bibinfo{author}{Gong, M.}, \bibinfo{year}{2014}.
\newblock \bibinfo{title}{Local metric learning for exemplar-based object
  detection}.
\newblock \bibinfo{journal}{IEEE Transactions on Circuits and Systems for Video
  Technology} \bibinfo{volume}{24}, \bibinfo{pages}{1265--1276}.
\bibitem[{Yu et~al.(2018)Yu, Zhao, Zheng, Zhang and You}]{yu2018hierarchical}
\bibinfo{author}{Yu, C.}, \bibinfo{author}{Zhao, X.}, \bibinfo{author}{Zheng,
  Q.}, \bibinfo{author}{Zhang, P.}, \bibinfo{author}{You, X.},
  \bibinfo{year}{2018}.
\newblock \bibinfo{title}{Hierarchical bilinear pooling for fine-grained visual
  recognition}, in: \bibinfo{booktitle}{Proceedings of the European Conference
  on Computer Vision (ECCV)}, pp. \bibinfo{pages}{574--589}.
\bibitem[{Zhou et~al.(2015)Zhou, Ni, Hong, Wang and Tian}]{zhou2015}
\bibinfo{author}{Zhou, Y.}, \bibinfo{author}{Ni, B.}, \bibinfo{author}{Hong,
  R.}, \bibinfo{author}{Wang, M.}, \bibinfo{author}{Tian, Q.},
  \bibinfo{year}{2015}.
\newblock \bibinfo{title}{Interaction part mining: A mid-level approach for
  fine-grained action recognition}, in: \bibinfo{booktitle}{Proceedings of the
  IEEE conference on computer vision and pattern recognition}, pp.
  \bibinfo{pages}{3323--3331}.
\bibitem[{Zhou et~al.(2014)Zhou, Ni, Yan, Moulin and Tian}]{zhou2014}
\bibinfo{author}{Zhou, Y.}, \bibinfo{author}{Ni, B.}, \bibinfo{author}{Yan,
  S.}, \bibinfo{author}{Moulin, P.}, \bibinfo{author}{Tian, Q.},
  \bibinfo{year}{2014}.
\newblock \bibinfo{title}{Pipelining localized semantic features for
  fine-grained action recognition}, in: \bibinfo{booktitle}{European conference
  on computer vision}, \bibinfo{organization}{Springer}. pp.
  \bibinfo{pages}{481--496}.
\bibitem[{Zhu et~al.(2018)Zhu, Jing, You, Zhang and Zhang}]{zhu2018video}
\bibinfo{author}{Zhu, X.}, \bibinfo{author}{Jing, X.Y.}, \bibinfo{author}{You,
  X.}, \bibinfo{author}{Zhang, X.}, \bibinfo{author}{Zhang, T.},
  \bibinfo{year}{2018}.
\newblock \bibinfo{title}{Video-based person re-identification by
  simultaneously learning intra-video and inter-video distance metrics}.
\newblock \bibinfo{journal}{IEEE Transactions on Image Processing}
  \bibinfo{volume}{27}, \bibinfo{pages}{5683--5695}.

\end{thebibliography}

\end{document}